\documentclass[11pt,a4paper]{article}
\usepackage[hyperref]{acl2017}
\usepackage{times}
\usepackage{latexsym}

\usepackage{graphicx}
\usepackage{url}
\usepackage{hyperref}
\usepackage{amsmath}
\usepackage{amsfonts}
\usepackage{enumitem}
\usepackage{multirow}

\usepackage[normalem]{ulem}
\usepackage{todonotes}

\aclfinalcopy 
\def\aclpaperid{224} 

\title{Understanding Neural Machine Translation by Simplification: \\The Case of Encoder-free Models}

\author{Gongbo Tang$^1$\quad Rico Sennrich$^{2,3}$\quad Joakim Nivre$^1$ \medskip\\
  $^1$Department of Linguistics and Philology, Uppsala University\\
  $^2$School of Informatics, University of Edinburgh\\
  $^3$Institute of Computational Linguistics, University of Zurich\\
  {\tt firstname.lastname@\{lingfil.uu.se, ed.ac.uk\}}}

\date{}

\begin{document}
\maketitle

\begin{abstract}
  In this paper, we try to understand neural machine translation (NMT) via simplifying NMT architectures and training encoder-free NMT models. In an encoder-free model, the sums of word embeddings and positional embeddings represent the source. The decoder is a standard Transformer or recurrent neural network that directly attends to embeddings via attention mechanisms. Experimental results show (1) that the attention mechanism in encoder-free models acts as a strong feature extractor, (2) that the word embeddings in encoder-free models are competitive to those in conventional models, (3) that non-contextualized source representations lead to a big performance drop, and (4) that encoder-free models have different effects on alignment quality for German$\rightarrow$English and Chinese$\rightarrow$English. 

\end{abstract}

\section{Introduction}

Neural machine translation (NMT) \cite{kal2013recurrent,sutskever2014sequence,bahdanau15joint,luong2015effective} has emerged in the last few years and has achieved new state-of-the-art performance. However, NMT models are black boxes for humans and are hard to interpret. NMT models employ encoder-decoder architectures where an encoder encodes source-side sentences and an attentional decoder generates target-side sentences based on the outputs of the encoder.
In this paper, we attempt to obtain a more interpretable NMT model by simplifying the encoder-decoder architecture. 
We train encoder-free models where the sums of word embeddings and sinusoid embeddings \cite{vaswani2017Attention} represent the source. The decoder is a standard Transformer \cite{vaswani2017Attention} or recurrent neural network (RNN) that attends to embeddings via attention mechanisms. 

As motivation for our architecture simplification, consider the attention mechanism\footnote{We refer to the encoder-decoder attention mechanism unless otherwise specified.} 
\cite{bahdanau15joint,luong2015effective}, which has been introduced to extract features from the hidden representations in encoders dynamically.
Attention and alignment were initially used interchangeably, but it was soon discovered that the attention mechanism can behave very differently from traditional word alignment \cite[see][]{ghader2017what,koehn2017challenges}.
One reason for this discrepancy is that the attention mechanism operates on representations that potentially includes information from the whole sentence due to the encoder's recurrent or self-attentional architecture.
Intuitively, bypassing these encoder layers and attending word embeddings directly could lead to a more alignment-like, and thus predictable and interpretable behavior of the attention model.

By comparing encoder-free models with conventional models, we can better understand the working mechanism of NMT, figure out which components are more crucial, and learn lessons for improvement. Experimental results show that there is a significant gap between the two models. We focus on exploring what leads to the big gap. 

As the embeddings in encoder-free Transformers (\textit{Trans-noEnc}) are only influenced by attention mechanisms, without the help of encoders, we hypothesize that the quality of embeddings leads to the gap between Transformers and \textit{Trans-noEnc} models. Thus we conduct both qualitative and quantitative evaluations of the embeddings from Transformers and \textit{Trans-noEnc} models. We also hypothesize that the attention distribution in \textit{Trans-noEnc} is not spread out enough for extracting contextual features. 
However, we find that word embeddings and attention distributions are not the major reasons causing the distinct gap. We further explore NMT encoders. We find that even NMT models with one layer encoder get significant improvement compared to encoder-free models which indicates that non-contextualized source representations lead to the evident gap. 

In encoder-free models, the attention attends to source embeddings rather than hidden representations fused with the context. We hypothesize that encoder-free models generate better alignments than default models. We evaluate the alignments generated on German$\rightarrow$English (DE$\rightarrow$EN) and Chinese$\rightarrow$English (ZH$\rightarrow$EN). We find that encoder-free models improve the alignments for DE$\rightarrow$EN but worsen the alignments for ZH$\rightarrow$EN.

\section{Related Work}

\subsection{Understanding NMT}

The attention mechanism has been introduced as a way to learn an alignment between the source and target text, and improves encoder-decoder models significantly, while also providing a way to interpret the inner workings of NMT models.
However, \newcite{ghader2017what} and \newcite{koehn2017challenges} have shown that the attention mechanism is different from a word alignment. While there are linguistically plausible explanations in some cases -- when translating a verb, knowledge about the subject, object etc.\ may be relevant information -- other cases are harder to explain, such as an off-by-one mismatch between attention and word alignment for some models. We suspect that such a pattern can be learned if relevant information is passed to neighboring representations via recurrent or self-attentional connections.

\newcite{ding2017visualizing} show that only using attention is not sufficient for deep interpretation and propose to use layer-wise relevance propagation to better understand NMT. 
\newcite{wang2018neuralHMM} replace the attention model with an alignment model and a lexical model to make NMT models more interpretable. The proposed model is not superior but on a par with the attentional model. They clarify the difference between alignment models and attention models by saying that that the alignment model is to identify translation equivalents while the attention model is to predict the next target word. 

In this paper, we try to understand NMT by simplifying the model. We explore the importance of different NMT components and what causes the performance gap after model simplification.

\subsection{Alignments and Source Embeddings}

\newcite{Nguyen2018improving} introduce a lexical model to generate a target word directly based on the source words. With the lexical model, NMT models generate better alignments. 
\newcite{Kuang2018attention} propose three different methods to bridge source and target word embeddings. 
The bridging methods can significantly improve the translation quality. Moreover, the word alignments generated by the model are improved as well.

Our encoder-free model is a simplification and only attends to the source word embeddings. We aim to interpret NMT models rather than pursuing better performance. 

Different from previous work, \newcite{zenkel2019adding} introduce a separate alignment layer directly optimizing the word alignment. The alignment layer is an attention network learning to attend to source tokens given a target token. The attention network can attend to either the word embeddings or the hidden representations or both of them. The proposed model significantly improves the alignment quality and performs as well as the aligners based on traditional IBM models.

\section{Experiments}

In addition to training Transformer and \textit{Trans-noEnc} models, we also compare \textit{Trans-noEnc} with NMT models based on RNNs (\textit{RNNS2S}). We train \textit{RNNS2S} models without encoders (\textit{RNNS2S-noEnc}), without attention mechanisms (\textit{RNNS2S-noAtt}), and without both encoders and attention mechanisms (\textit{RNNS2S-noAtt-noEnc}) to explore which component is more important for NMT. We also investigate the importance of positional embeddings in \textit{Trans-noEnc}. 

\subsection{Experimental Settings}
We use the \textit{Sockeye} \cite{Hieber2017sockeye} toolkit, which is based on MXNet \cite{chen2015mxnet}, to train models. Each encoder/decoder has 6 layers. For \textit{RNNS2S}, we choose long short-term memory (LSTM) RNN units. Transformers have 8 attention heads. The size of embeddings and hidden states is 768. We tie the source, target, and output embeddings. The dropout rate of embeddings and Transformer blocks is set to 0.1. The dropout rate of RNNs is 0.2. 
All the models are trained with a single GPU. During training, each mini-batch contains 2,048 tokens. A model checkpoint is saved every 1,000 updates. We use \textit{Adam} \cite{Kingma2014AdamAM} as the optimizer. The initial learning rate is set to 0.0001. If the performance on the validation set has not improved for 8 checkpoints, the learning rate is multiplied by 0.7. We set the early stopping patience to 32 checkpoints. 

The training data is from the WMT15 shared task \cite{wmt15} on Finnish--English (FI--EN). We choose \textit{newsdev2015} as the validation set and use \textit{newstest2015} as the test set. All the BLEU \cite{papineni2002} scores are measured by \textit{SacreBLEU} \cite{matt2018sacrebleu}. There are about 2.1M sentence pairs in the training set after preprocessing. We learn a joint BPE model with 32K subword units \cite{sennrich16sub}. We employ the models that have the best perplexity on the validation set for the evaluation. We set the beam size to 8 during inference.

To test the universality of our findings, we conduct experiments on DE$\rightarrow$EN and ZH$\rightarrow$EN as well. For DE$\rightarrow$EN, we use the training data from the WMT17 shared task \cite{wmt17}. We use \textit{newstest2013} as the validation set and \textit{newstest2017} as the test set. We learn a joint BPE model with 32k subword units. For ZH$\rightarrow$EN, we choose the CWMT parallel data of the WMT17 shared task for training. We use \textit{newsdev2017} as the validation set and \textit{newstest2017} as the test set. We apply Jieba\footnote{\url{https://github.com/fxsjy/jieba}} to Chinese segmentation. We then learn 60K subword units for Chinese and English separately. There are about 5.9M and 9M sentence pairs in the training set after preprocessing in DE$\rightarrow$EN and ZH$\rightarrow$EN, respectively.

\subsection{Results}
Table \ref{table-result} shows the performance of all the trained models. Encoder-free models (\textit{NMT-noEnc}s) perform rather poorly compared to conventional NMT models.\footnote{We also trained a \textit{Transformer} with less parameters (64.3M). The \textit{Transformer} still achieved a significantly better BLEU score (18.2) than \textit{Trans-noEnc} which means that the number of parameters is not the primary factor in this case.}
It is interesting that \textit{Trans-noEnc} obtains a BLEU score similar to the \textit{RNNS2S} model. Even though the attention networks only attend to the non-contextualized word embeddings, \textit{Trans-noEnc} still performs as well as the \textit{RNNS2S} by paying attention to the context with multiple attention layers. \newcite{Tang2018why} find that the superiority of Transformer models is attributed to the self-attention network which is a powerful semantic feature extractor. Given our results, we conclude that the attention mechanism is also a strong feature extractor in \textit{Trans-noEnc} without self-attention in the encoder. 

\begin{table}[htbp]
\begin{center}
\scalebox{0.9}{
\begin{tabular}{|l|c|c|c|}
\hline Model&Param.&PPL&BLEU\\
\hline \textit{Transformer}&104.4M&\phantom{0}9.6& 18.9\\
\textit{Trans-noEnc}&\phantom{0}71.4M&11.7& 15.9\\
\hline \textit{RNNS2S}& \phantom{0}91.5M&14.9& 15.9\\
\textit{RNNS2S-noEnc}&\phantom{0}64.3M&25.2 &12.5 \\
\textit{RNNS2S-noAtt}&\phantom{0}90.3M& 33.3& \phantom{0}8.2\\
\textit{RNNS2S-noAtt-noEnc}&\phantom{0}63.1M& 53.7& \phantom{0}4.1\\
\hline \textit{Trans-noEnc-noPos}&\phantom{0}71.4M&26.6& \phantom{0}7.1\\
\hline
\end{tabular}}
\caption{\label{table-result} The performance of NMT models. PPL is the perplexity on the development set. BLEU scores are evaluated on \textit{newstest2015}. ``Param.'' denotes the number of parameters. }
\end{center}
\end{table}

\begin{table*}[htbp]
\begin{center}
\begin{tabular}{|c|c|c|}
\hline \multirow{2}{*}{Word} & \multicolumn{2}{|c|}{Neighbors} \\
\cline{2-3} &\textit{Transformer}&\textit{Trans-noEnc}\\
\hline  more& less, better, greater, most, \textbf{further}& less, greater, better, \textbf{fewer}, most \\
\hline  for& to, in, on, of, \textbf{with}& to, in, of, on, \textbf{towards} \\
\hline ole (not)& 
\begin{tabular}{@{}c@{}}olekaan (not the), \textbf{kykene} (unable to), kuulu\\(part of), \textbf{pysty} (upright), \textbf{ollut} (been) \end{tabular}  & \begin{tabular}{@{}c@{}} olekaan, kuulu (part of), \textbf{ei}\\(no/not), \textbf{ene} (a suffix), \textbf{liity} (sign up) \end{tabular}  \\
\hline \begin{tabular}{@{}c@{}}Arvoisa\\ (honorable) \end{tabular}  & 
\begin{tabular}{@{}c@{}}arvoisa, Arvoisat (honorable), \textbf{arvoisaa}, \\ \textbf{arvoisan} (honorable), hyv\"{a}t (honorable) \end{tabular}  & \begin{tabular}{@{}c@{}} arvoisa, \textbf{arvoisat}, hyv\"{a}t,\\ Arvoisat, \textbf{Hyv\"{a}} (honorable) \end{tabular}  \\
\hline
\end{tabular}
\caption{\label{table-embeddings-diff} Neighbors of the selected word embeddings. Bold words are distinct neighbors.}
\end{center}
\end{table*}

\noindent
The attention mechanism improves encoder-decoder architectures significantly. However, there are no empirical results to clarify whether encoders or attention mechanisms are more important for NMT models. We compare \textit{RNNS2S-noAtt}, \textit{RNNS2S-noEnc}, and \textit{RNNS2S-noAtt-noEnc} to explore which component contributes more to NMT models.\footnote{Because the encoders and decoders in Transformers are only connected via attention, we only conduct this experiment on \textit{RNNS2S} models.} 
In Table \ref{table-result}, \textit{RNNS2S-noEnc} performs much better than \textit{RNNS2S-noAtt}. Moreover, the gap between \textit{RNNS2S-noEnc} and \textit{RNNS2S-noAtt-noEnc} is distinctly larger than the gap between \textit{RNNS2S-noAtt} and \textit{RNNS2S-noAtt-noEnc}. These results hint that attention mechanisms are more powerful than encoders in NMT. 

The positional embedding is also very important to Transformers which holds the sequential information. We are interested in the extent to which the positional embedding affects the translation performance. We further simplify the model by removing the positional embedding in the source (\textit{Trans-noEnc-noPos}). \textit{Trans-noEnc-noPos} has a dramatic drop in BLEU score. It is even worse than \textit{RNNS2S-noAtt}. This result indicates that positional information is indeed crucial for Transformers.

\section{Analysis}

\textit{Trans-noEnc} is obviously inferior to \textit{Transformer} but we are more interested in investigating what causes the performance gap. In this section, we will test our hypotheses on embedding quality and attention distributions. 

\subsection{Embeddings} 
\label{sub:embeddings}

Word embeddings are randomly initialized by default and learned during training. As the embeddings in \textit{Trans-noEnc} are only updated by attention mechanisms, we hypothesize that embeddings in \textit{Trans-noEnc} are not well learned and therefore affect translation performance. We test our hypothesis by (1) evaluating the embeddings in the two models manually and (2) initializing \textit{Trans-noEnc} with the learned embeddings in \textit{Transformer} as pre-trained embeddings. 

\paragraph{Qualitative Evaluation}
We select the 150 most frequent tokens from the vocabulary and then manually evaluate the quality of embeddings by comparing the 5 nearest neighbors. 

The quality of English word embeddings is quite good based on the output of neighbors. Finnish word embeddings are not as good as English word embeddings. Table \ref{table-embeddings-diff} exhibits four examples, two English words, ``more'', ``for'' and two Finnish word, ``ole'' (not), ``Arvoisa'' (honorable). The neighbors of ``more'' in \textit{Transformer} and \textit{Trans-noEnc} are all quite related words, including comparatives and ``most'' which is the superlative of ``more''. The words ``further'' and ``fewer'' are more different neighbors but both are related to ``more''. 
For the Finnish word ``ole'' (not), both models have negative words as neighbors, but there are different unrelated words as well. We can see that the qualities of neighbors in two embedding matrices are close. We cannot easily distinguish which embedding matrix is better based on the neighbors. 

\paragraph{Quantitative Evaluation}
In addition to the qualitative evaluation, we also conduct a quantitative evaluation. We first employ the learned embeddings from \textit{Transformer} to initialize the embedding parameters in \textit{Trans-noEnc}. The pre-trained embeddings can be either fixed or not fixed during training. 
Table \ref{table-embeddings} gives the BLEU scores of these models. The pre-trained embeddings slightly improve the BLEU score.

\begin{table}[htbp]
\begin{center}
\scalebox{0.9}{
\begin{tabular}{|c|c|c|c|}
\hline Embeddings&Random&Fixed&Not-fixed\\
\hline BLEU&15.9&16.1& 16.2\\
\hline
\end{tabular}}
\caption{\label{table-embeddings} BLEU scores of \textit{Trans-noEnc}s with different embedding initialization. ``Random'' means no pre-trained embeddings. ``Fixed'' and ``Not-fixed'' denote using pre-trained embeddings. } 
\end{center}
\end{table}

\noindent
The evaluation reveals that the embeddings from \textit{Trans-noEnc} are competitive to those of \textit{Transformer}.
Thus, we can rule out differences in embedding quality as the main factor for the performance drop.

\subsection{Attention Distribution} 
\label{sub:attention_distribution}

The attention networks in \textit{Trans-noEnc} only attend to word embeddings. To better capture the sentence-level context, the attention networks need to distribute more attention to the context. We test our hypothesis that the attention distributions in \textit{Trans-noEnc} are not as distributed as those in \textit{Transformer}. If the attention distributions in \textit{Transformer} are more spread out than those in \textit{Trans-noEnc}, it means that smaller weights are distributed to contextual features by \textit{Trans-noEnc}. 
\begin{equation} \label{eq:attention-entropy}
E_{At}(y_{t}) = - \sum_{i=1}^{|x|} At(x_{i},y_{t}) \log At(x_{i},y_{t}) 
\end{equation} 
We use attention entropy (Equation \ref{eq:attention-entropy}) \cite{ghader2017what} to measure the concentration of the attention distribution at timestep $t$. We then average the attention entropy at all the timesteps as the final attention entropy. $x_{i}$ denotes the $i$th source token, $y_{t}$ is the prediction at timestep $t$, and $At(x_{i},y_{t})$ represents the attention distribution at timestep $t$. The attention mechanism in Transformer has multiple layers, and each layer has multiple heads. In each layer, we average the attention weights from all the heads.  

\begin{figure}[htbp]
\centering
        \includegraphics[totalheight=4.1cm]{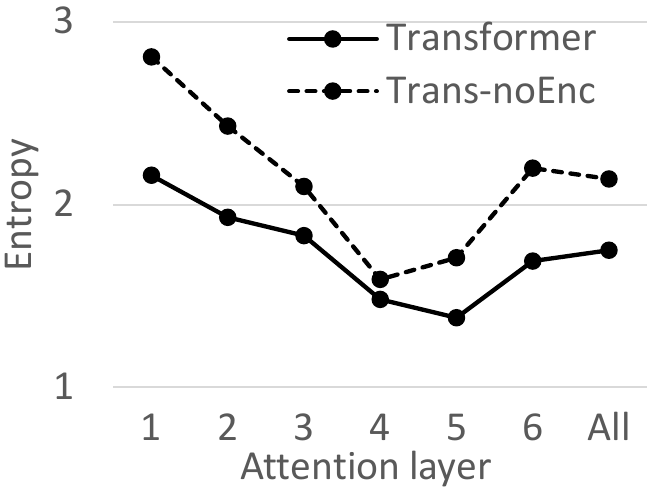}
    \caption{The attention entropy of each attention layer and the entire attention mechanism.}
    \label{fig:entropy}
\end{figure}

Figure \ref{fig:entropy} shows the entropy of attention distributions in both models. The attention distributions are consistent with the finding in \newcite{Tang2018WSD} that the distribution gets concentrated first and then becomes distributed again. \textit{Transformer} has lower entropy, which potentially is because the contextual information has been encoded in the hidden representations.  The attention entropy of \textit{Trans-noEnc} is clearly higher than that of \textit{Transformer} in each attention layer. The attention in \textit{Trans-noEnc} tends to extract features from source tokens more uniformly which indicates that the attention mechanism compensates for the fact that embeddings are non-contextualized by distributing attention across more tokens.

\subsection{Encoders}
\label{sub:encoders}

We have shown that embeddings and attention distributions are not the primary reasons causing the gap between \textit{Transformer} and \textit{Trans-noEnc}. Therefore, we move to explore encoders. 

Encoders are responsible for providing source hidden representations to the decoder. Encoder-free models have to use word embeddings to represent source tokens without the help of encoders. Thus, the source-side representations probably lead to the performance gap. 

We train NMT models with different encoder layers. Table \ref{table-encoder-layer} displays the performance of Transformer models that have different layers in the encoder. It is clear that even the model with only a 1-layer encoder outperforms \textit{Trans-noEnc} (0-layer) by 1.7 BLEU points, which accounts for 56.7\% of the performance gap. The results seem to show that source-side hidden representations are crucial in NMT. 

\begin{table}[htbp]
\begin{center}
\begin{tabular}{|c|c|c|c|}
\hline Layers&Param.&PPL&BLEU\\
\hline 0&\phantom{0}71.4M&11.7& 15.9\\
\hline 1&\phantom{0}76.9M&10.3& 17.6\\
\hline 3& \phantom{0}87.9M&\phantom{0}9.9& 18.4\\
\hline 5&\phantom{0}98.9M&\phantom{0}9.5 &18.6 \\
\hline 6&104.4M& \phantom{0}9.6& 18.9\\
\hline
\end{tabular}
\caption{\label{table-encoder-layer} The performance of Transformer models that have different layers in the encoder, including the perplexity (PPL) on the development set and the BLEU scores on \textit{newstest2015}. }
\end{center}
\end{table}

\noindent
It has been shown that encoders could extract syntactic and semantic features in NMT \cite{belinkov2017what,belinkov2017evaluating,Poliak2018on}. 
In the meantime, contextual information is encoded in hidden representations as well. Hence we conclude that the quality of source representations is the main factor causing the big gap between \textit{Transformer} and \textit{Trans-noEnc}. 

In Table \ref{table-result-language}, our additional experiments on DE$\rightarrow$EN and ZH$\rightarrow$EN confirm that models with contextualized representations are much better. Transformer models always outperform \textit{Trans-noEnc} models substantially. 

\begin{table}[htbp]
\begin{center}
\scalebox{0.9}{
\begin{tabular}{|c|c|c|c|c|}
\hline Lan.&\textit{Trans-noEnc}&\textit{Transformer}&Impr.\\
\hline DE$\rightarrow$EN&29.5&32.6&10.5\%\\
\hline ZH$\rightarrow$EN&18.5&20.9&13.0\%\\
\hline
\end{tabular}}
\caption{\label{table-result-language} The improvement (Impr.) of employing encoders in \textit{Trans-noEnc}s on DE$\rightarrow$EN and ZH$\rightarrow$EN. }
\end{center}
\end{table}

\section{Alignment} 
\label{sec:alignment}

The weights of the attention mechanism can be interpreted as an alignment between the source and target text.
We further explore whether encoder-free models have better alignments than default models. 
We evaluate the alignments on two manually annotated alignment data sets. The first one has been provided by RWTH,\footnote{\url{https://www-i6.informatik.rwth-aachen.de/goldAlignment/}} and consists of 508 DE$\rightarrow$EN sentence pairs. The other one is from \newcite{liu2015contrastive} and contains 900 ZH$\rightarrow$EN sentence pairs. We apply alignment error rate (AER) \cite{Och2003comparison} as the evaluation metric. 

Following \newcite{luong2015effective,Kuang2018attention}, we also force the models to produce the reference target words during inference to get the alignment between input sentences and their reference outputs. We merge the subwords after translation following the method in \newcite{koehn2017challenges}.\footnote{(1) If an input word is split into subwords, we sum their attention weights. (2) If a target word is split into subwords, we average their attention weights.} 
We sum the attention weights in all attention heads in each attention layer.\footnote{Following \newcite{Tang2018WSD}, we tried maximizing the attention weights as well but got worse alignment quality.} 
Given a target token, the source token with the highest attention weight is viewed as the alignment of the current target token \cite{luong2015effective}. However, a source token maybe aligned to multiple target tokens and vice versa. Therefore, we also align a source token to the target token that has the highest attention weight given the source token. Experimental results show that the bidirectional method achieves higher alignment quality. 

\begin{figure}[htbp]
\centering
        \includegraphics[totalheight=3.8cm]{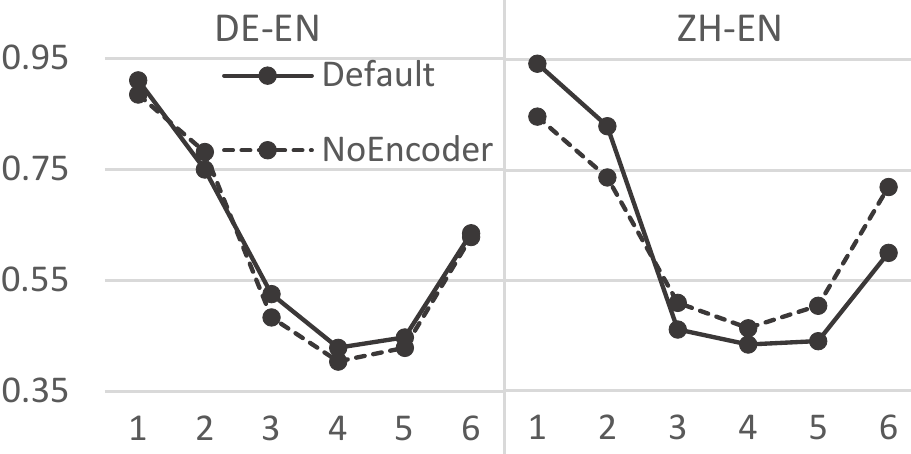}
    \caption{The AER scores of alignments in different attention layers on DE$\rightarrow$EN and ZH$\rightarrow$EN.}
    \label{fig:align-deen}
\end{figure}

Figure \ref{fig:align-deen} displays the evaluation results. The alignment in the fourth attention layer achieves the best performance. Therefore, we only compare the alignments in the fourth layer. In DE$\rightarrow$EN, the encoder-free model has a lower AER score (0.41) than the default model (0.43) which accords with our hypothesis. However, in ZH$\rightarrow$EN, the alignment quality of the encoder-free model (0.46) is worse than that of the default model (0.43). The effect on alignment quality is not clear-cut for encoder-free models given limited language pairs. 

\section{Conclusion}
\label{sec:conclusion}

To better understand NMT, we simplify the attentional encoder-decoder architecture by training encoder-free NMT models in this paper. 
The non-contextualized source representations in encoder-free models cause a big performance drop,
but the word embeddings in encoder-free models are shown competitive to those in default models.
Also, we find that the attention component in encoder-free models is a powerful feature extractor, and can partially compensate for the lack of contextualized encoder representations.

Regarding the interpretability of attention, our results do not show that the attention mechanism in encoder-free models is consistently more alignment-like:
only attending to source embeddings improves the alignment quality on DE$\rightarrow$EN but makes the alignment quality worse on ZH$\rightarrow$EN. 

\section*{Acknowledgments}
We thank all reviewers for their valuable and insightful comments. 
Gongbo Tang is mainly funded by the Chinese Scholarship Council (grant number \texttt{201607110016}). 

\bibliography{noencoder}
\bibliographystyle{acl_natbib}

\end{document}